\documentclass{article}
\usepackage[numbers]{natbib}
\usepackage {arxiv}

\usepackage{cite}
\usepackage{float}
\usepackage{amsmath,amssymb,amsfonts}
\usepackage{algorithmic}
\usepackage{graphicx}
\usepackage{textcomp}
\usepackage{booktabs}
\usepackage{amsmath}
\usepackage{subfigure}
\usepackage[table,xcdraw]{xcolor}
\usepackage{soul}
\usepackage[dvipsnames]{xcolor}
\usepackage{hyperref}
\hypersetup{hidelinks}

\def\BibTeX{{\rm B\kern-.05em{\sc i\kern-.025em b}\kern-.08em
    T\kern-.1667em\lower.7ex\hbox{E}\kern-.125emX}}

\title{\LARGE \bf
Importance Scoring of Transformer Attention Heads in Learning Tabular Data}

\author{Ahmad JadAllah, ~
        Kazi F. Akhter\\
Department of Computer Science\\
Tennessee State University\\
Nashville, TN, USA\\
\and 
{\bf Md. Kamrozzaman Bhuiyan}\\
Enosis Solutions\\
Dhaka, Bangladesh\\
\and
 {\bf Manar D. Samad}\\
 Department of Computer Science\\
 North Carolina Agricultural and Technical University\\
 Greensboro, NC, USA
}

\begin{document}

\maketitle

\begin{abstract}

Computationally demanding and opaque deep learning models can be better understood and optimized by analyzing how they transform data. While deep transformers have been widely studied in computer vision and natural language processing, their application in tabular data remains relatively underexplored. This paper presents one of the first applications of an importance-scoring metric to interpret multi-head transformer models in learning from tabular data. Experiments conducted on 40 diverse tabular datasets demonstrate robustness to head drops based on the proposed head importance score. In 72.5\% of the experimental examples, the model remains most resilient to performance drop when heads with the lowest importance scores are gradually decreased. In contrast, removing the most important attention head first results in the greatest reduction in classification performance. A closer look at individual head importance scores across six attention layers reveals that important heads are scattered across layers without consistent layer-specific trends. In contrast to the image and language domains, the importance of individual attention heads varies considerably across tabular datasets that differ in their schemas and feature spaces. The proposed importance score can improve efficiency and redundancy within transformer architectures. The source code for measuring the importance of individual attention heads is made publicly available. ~\footnote{\href{https://github.com/mdsamad001/Attention-Head-Importance-Scoring-for-Tabular-Data.git} {https://github.com/mdsamad001/Attention-Head-Importance-Scoring-for-Tabular-Data.git}}. 
 \end{abstract}


\section{Introduction}

 Transformer models have been introduced primarily for language translation and decoding~\citep{vaswani2017_attention, radford2019_gpt2}, with applications that extend to computer vision~\citep{dosovitskiy2020_ViT}. Extensive research has been conducted to evaluate the efficacy of transformers in language and vision applications, especially those involving image and text data. In contrast, tabular data, structured in rows and columns, have seen limited research involving transformer models. Tabular data, which are prevalent in enterprise relational databases and spreadsheets across numerous application domains, are inherently distinct from the commonly studied text and image data. While images and text are sequential and homogeneous distributions of pixels and words, respectively, tabular data are a permutation-invariant collection of heterogeneous features~\citep{rabbani2024_survey, Grinsztajn2022_survey, Borisov2022_survey}. As a result, numerous intuitions about deep learning that come from language and vision work are challenged when applied to tabular data. These structural differences among tabular datasets may cause individual attention heads to contribute differently across datasets, rather than following the hierarchical patterns observed in language and vision models.

 Indeed, transformer models have seen impressive success in classifying~\citep{hollmann2025_tabpfnv2, somepalli2022_saint, kossen2021_npt}, transfer learning~\citep{wang2022_transtab, kowsar2025_lattle}, and the imputation of missing values in tabular datasets~\citep{kowsar2025_deepifsac}. In most cases, transformer architectures developed for large language models are directly adapted to learning tabular data without optimizing for a performance-computational cost trade-off. Weight pruning methods for optimizing transformers have been investigated using language data~\citep{michel2019_sixteenhead}, which have not been explored well in the context of tabular data. Although extensive language corpora exist for training deep transformer models, such large-scale architectures are often not suitable for many tabular datasets with only a limited number of samples. This paper is among the first to quantify the contribution of individual attention heads of a deep transformer model in learning tabular data across different domains and scales. Measuring the contribution of each individual attention head supports selecting and pruning heads for model compression, and this process, in turn, yields an explainable deep learning framework for tabular transformer models.

 The remainder of the paper is organized as follows. Section~\ref{relatedwork} provides background information and a brief review of the literature. Section~\ref{methods} outlines the methodological framework for the head importance scoring metric. The experimental cases and evaluation methods are discussed in Section~\ref{experiment}. Section~\ref{results} presents the key findings and their discussion. The paper concludes in Section~\ref{conclu}.

\section{Related work} \label{relatedwork}


Multiple attention heads enable a transformer to simultaneously capture diverse patterns from the input, each head attending to different aspects of the input data~\citep{vaswani2017_attention}.
Quantifying the contribution of individual heads within an attention layer is central to understanding and controlling redundancy in multi-head transformer architectures. Transformers with multi-head attention have primarily been designed and extensively investigated to model language data. By contrast, because tabular data are heterogeneous and permutation-invariant in their feature space, they may engage with attention heads in a fundamentally different way than semantic word sequences in language data.

Transformer architectures have traditionally been configured as a rigid black box with a fixed number of attention layers and heads per layer, regardless of the target data domain or data types~\citep{vaswani2017_attention, devlin2019_bert}. For example, the first proposed transformer architecture~\citep{vaswani2017_attention} uses six attention layers with eight attention heads per layer. The language encoder model, BERT~\citep{devlin2019_bert}, has 12 layers with 12 heads per layer. Similar setups are routinely adopted across numerous application areas, thereafter, without further tuning to the particular task or dataset. There has been no standard approach for selecting or justifying the number of transformer layers and attention heads. Previous work has demonstrated that these standard transformer architectures are largely redundant in domain-specific applications~\citep{bao2026_redundant}. Therefore, tailoring transformer architectures to specific datasets and schemas can improve performance, increase computational efficiency, and offer more transparent interpretations of otherwise opaque deep learning mechanisms. 

This paper draws its inspiration from a related approach previously introduced to study redundancy in transformer-based language models. Michel et al.~\citep{michel2019_sixteenhead} have introduced a head-importance metric $I_h$ based on the sensitivity of the loss to the activation of individual heads, which is  
used as a post-hoc criterion to identify and remove redundant heads from an 
already-trained model. However, the authors do not share the source code for their implementation of the head importance metric. To the best of our knowledge, similar studies have not been performed on tabular datasets with heterogeneous feature spaces. Many application domains provide datasets with limited sample sizes, which may not fully leverage the strength of deep learning methods.  

This paper is one of the first to investigate the head-importance scoring strategy for studying redundancy within a multi-head transformer architecture when learning from tabular datasets across varying domains and schemas. The importance scores for the head are used to dynamically tune head activation during training. The final scores are used to drop heads cumulatively to interpret the importance of individual heads across multiple attention layers. We hypothesize that the vanilla transformer architecture is redundant for certain tabular datasets, which may benefit from data-specific model customization and provide useful interpretation of its contributions.

The contributions of this paper are as follows. First, we measure the importance of individual transformer attention heads for learning on tabular datasets with a heterogeneous feature space.  Second, because the authors of ~\citep{michel2019_sixteenhead} do not release their implementation, we implement the importance metric ourselves, adapting it to learning tabular data, and have made our code publicly available. Third, as an exploratory analysis, we examine a dynamic soft-tuning mechanism for attention head activation to study its effect across datasets. Fourth, we visualize the importance scores of individual heads across different attention layers to support the interpretation of the otherwise opaque transformer model. Fifth, we conduct experiments on 40 tabular datasets that span various domains and sizes to highlight heterogeneity in transformer interpretations.

\section{Methodology} \label{methods}

This section introduces metrics that quantify the importance of attention heads in transformer models, along with an activation-tuning strategy used during transformer training.

\subsection {Preliminaries on tabular data}

While images are spatial distributions of pixels and language is a sequential arrangement of words, tabular data are arranged in a fixed schema with rows and columns. Tabular data consist of $n$ samples organized in rows, where each row is described by $d$ feature columns, forming a data matrix $X \in \mathbb{R}^{n \times d}$. Unlike uniform image pixels or words, the feature space in tabular data is heterogeneous in both types and scales. The diversity of feature spaces is further increased by the variety of application domains, in which the feature space may combine sparse, ordinal, numerical, and categorical variables. Although vision and language models often exhibit stable performance and similar behavior across many datasets, these observations generally do not extend to tabular data due to the aforementioned heterogeneity. The distinctive nature of tabular data and the importance of incorporating data-specific inductive biases are not widely acknowledged in the deep learning literature. 

\subsection{Attention head importance score }

The head importance score measures the contribution of each attention head to the loss during training. Inspired by the work of Michel et al.~\citep{michel2019_sixteenhead} in the natural language domain, we obtain the importance score as the expected absolute inner product between head activation and the loss gradient with respect to that activation as follows.
\begin{equation}
I_h^{(l)} = \mathbb{E}_{x \sim X} \left| \mathrm{Att}_h^{(l)}(x)^\top \cdot \frac{\partial \mathcal{L}(x)}{\partial \mathrm{Att}_h^{(l)}(x)} \right|
\label{eq:head_importance}
\end{equation}

Here, $\mathrm{Att}_h(x) \in \mathbb{R}^{d \times d_h}$ is the activation output representation of the attention head $h$ for input $x$, where $d$ is the dimension of $x$ and $d_h$ is the dimension of the attention embedding. The expectation $\mathbb{E}_{x \sim X} $ is taken over the data distribution $X$. The absolute value skips the cancellation of opposite signs to reflect the magnitude of each head's influence on the loss regardless of direction~\citep{michel2019_sixteenhead}. In practice, the exact computation of this expectation $\mathbb{E}_{x \sim X} $ is infeasible. We therefore approximate $I_h$ at each training epoch $e$ by evaluating the importance over the full training set in mini-batches. Specifically, at epoch $e$, the training set is divided into mini-batches $B_1, \dots, B_K$. For each mini-batch $B_k$, a forward and backward pass is performed, after which the per-head output tensor $\mathrm{Att}_h^{(l)}(x) \in \mathbb{R}^{|B_k| \times d \times d_h}$ and its gradient $\frac{\partial \mathcal{L}}{\partial \mathrm{Att}_h^{(l)}(x)}$ of the same shape are available. The per-batch importance is computed as the absolute value of the element-wise product summed over the batch, sequence, and head-dimension axes. These per-batch scores are then accumulated across all mini-batches and divided by the total number of training samples $N$, producing the epoch-level estimate. The importance score of a head is obtained as the gradient-weighted response of attention heads. A higher gradient of training loss would increase the importance of the respective head.

\subsection{Soft dynamic tuning of attention heads}
We employ a continuous gating mechanism to dynamically regulate the activation of individual attention heads during training. For each head $h$ in layer $l$, a gate value $\{0 \leq \xi_h^{(l,e)} \leq 1\}$ scales the head activation before it is combined with the outputs of the remaining heads in that layer:
\begin{equation}
h_{\mathrm{scaled}}^{(l)}(x) = \xi_h^{(l,e)} \cdot \mathrm{Att}_h^{(l)}(x)
\label{eq:head_scaled}
\end{equation}
where $\mathrm{Att}_h^{(l)}(x)$ is the pre-gating head activation defined in Section~III-A. The scaled activations of all $K$ parallel heads in an attention layer are concatenated and projected through the output matrix.
After epoch $e$, the importance vector containing individual head importance scores is as follows.
\begin{equation}
\mathbf{I}^{(l,e)} = \left(I_1^{(l,e)}, \ldots, I_H^{(l,e)}\right).
\end{equation}
The gate value to tune the activation of each head ($h$) under layer $l$ is obtained as follows.
\begin{equation}
\xi_h^{(l,e)} = \frac{I_h^{(l,e)}}{\left\|\mathbf{I}^{(l,e)}\right\|_2 + \epsilon},
\label{eq:gate_update}
\end{equation}
where $\|\cdot\|_2$ denotes the Euclidean norm computed over the $K$ heads in layer $l$, and $\epsilon = 10^{-8}$ prevents division by zero. Training involving soft tuning proceeds in two phases. During an initial warmup period lasting $E_w$ epochs, a constant gate value (unity) is applied while updating only the model parameters.  This enables the network to develop stable feature representations before activating the importance score-based tuning mechanism. Following the warm-up phase, the importance scores $I_h^{(l,e)}$ are calculated over the entire training set using the method outlined in Section~III-A, and the gate values are then updated according to Equation~\eqref{eq:gate_update}.

\subsection{Transformer architecture}

We employ the TransTab architecture~\citep{wang2022_transtab}, a transformer model designed for tabular data, using six attention layers, each containing eight attention heads, yielding a total of 48 heads. In TransTab, the feedforward network is configured with a hidden dimension of 256, employs ReLU as its activation function, and incorporates dropout. For representing feature names and categorical values, the transformer relies on embeddings derived from a pre-trained BERT~\citep{devlin2019_bert} model. The categorical feature name (``color'') and its value (``red'') 
are each tokenized and passed through the pre-trained BERT model to obtain their respective embeddings, which are then concatenated to form the categorical feature embedding $E_\text{cat}$. For numerical features, the embedding of the feature name (``age'') is multiplied element-wise by 
the scalar feature value ($40$) to produce the numerical feature embedding $E_\text{num}$. The categorical and numerical embeddings are concatenated across all features of a sample to form a unified token sequence, which is passed as input to the transformer encoder.

\section{Experiments and evaluations} \label{experiment}

Table~\ref{tab:expt} shows eight experimental cases. The first two scenarios compare model training with and without soft tuning, resulting in fixed performance scores without removing any heads. The next two baselines compare setups with and without soft tuning, where heads are randomly dropped only after the model has finished training. The following four scenarios consider cumulative head drop after completion of model training, progressing from the least important head with \(I_{min}\) to the most important head with \(I_{max}\), and vice versa. As a practical test of a head’s importance, one would expect that eliminating the most important head should cause the greatest drop in performance, thereby offering a way to validate the $I_h$ metric. Intuitively, cumulative head removal is expected to show a gradual decline in downstream test accuracy. The cases that stand out as the strongest in downstream performance after head drop will be identified as the most resilient methods to head drops, facilitating the best performance-cost trade-off.

For model evaluation, datasets are divided into 70:20:10 ratios for training, validation, and test sets, respectively. The validation set is used only to select the best model during training, whereas the final evaluation is performed only on the left-out test data samples. We use the area under the receiver operating characteristic curve (AUC) as the evaluation metric. For multiclass datasets, the AUC is computed using the one-vs-rest (OVR) method with macro averaging. For experimental cases involving the removal of attention heads, we obtain test AUC scores as a function of the number of cumulative heads removed for individual datasets. The importance score is used to determine the order in cumulative head removal across six attention layers. We count the number of datasets in which an experimental case ranks best throughout progressive head removal. The number of datasets achieving the highest AUC scores for an experimental case is determined when 10, 20, 30, and 40 heads are successively removed following the head importance scores. We also visualize the importance scores of 48 attention heads distributed across six attention layers from the bottom to the top of the transformer model. The statistical difference between the experimental cases is obtained using Wilcoxon signed-rank test across the 40 datasets. Cohen’s d was also calculated to report the effect size.


\begin{table}[t]
    \centering
        \caption{Experimental scenarios designed to assess how effectively various attention-head soft-tuning and head-dropping strategies support learning on tabular datasets.}
    \begin{tabular}{lccr}
    \toprule 
         Case & Soft  & Head dropping& Comment \\
         &tuning &strategy & \\
        \midrule
      1&  No  & None & Fixed reference\\
        2&  Yes  & None& Fixed reference\\
        \midrule
          3 & No & Random & Baseline\\
         4& Yes & Random& Baseline\\
         \midrule
         5 & No & $I_{min}$ $\rightarrow$ $I_{max}$ &  Drop the least important first\\
         6& Yes & $I_{min}$ $\rightarrow$ $I_{max}$& Drop the least important first\\
        7 & No & $I_{max}$ $\rightarrow$ $I_{min}$  & Dropping the most important heads\\
         8& Yes & $I_{max}$ $\rightarrow$ $I_{min}$ & Dropping the most important heads\\
         \bottomrule
    \end{tabular}
    \label{tab:expt}
\end{table}

\begin{figure}[t]
\centering
\subfigure[Feature distribution] {\includegraphics[width=0.3\linewidth]{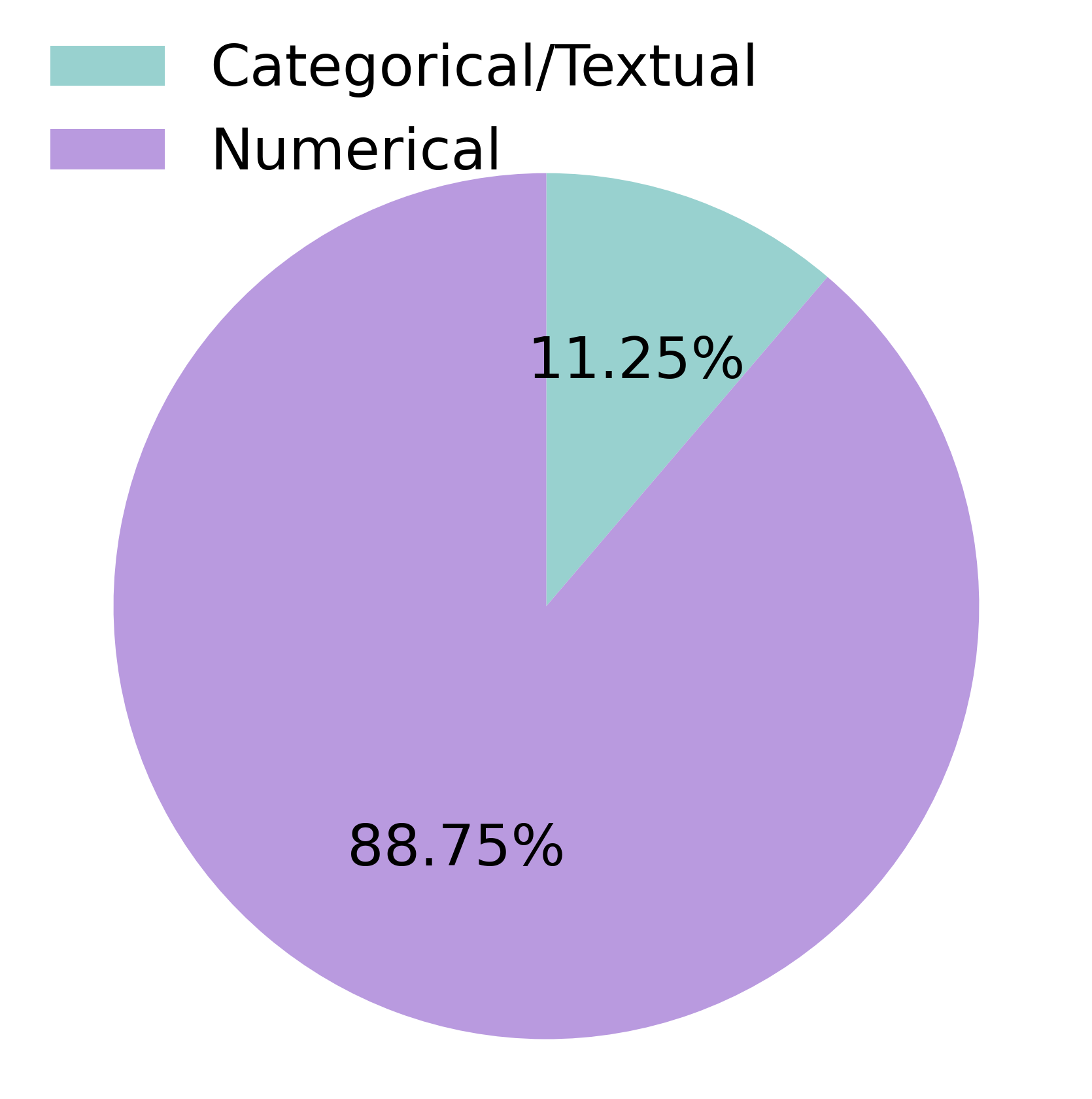}
\label{fig:four_a}}
\subfigure[Distribution of tabular datasets] { \includegraphics[width=0.6\linewidth]{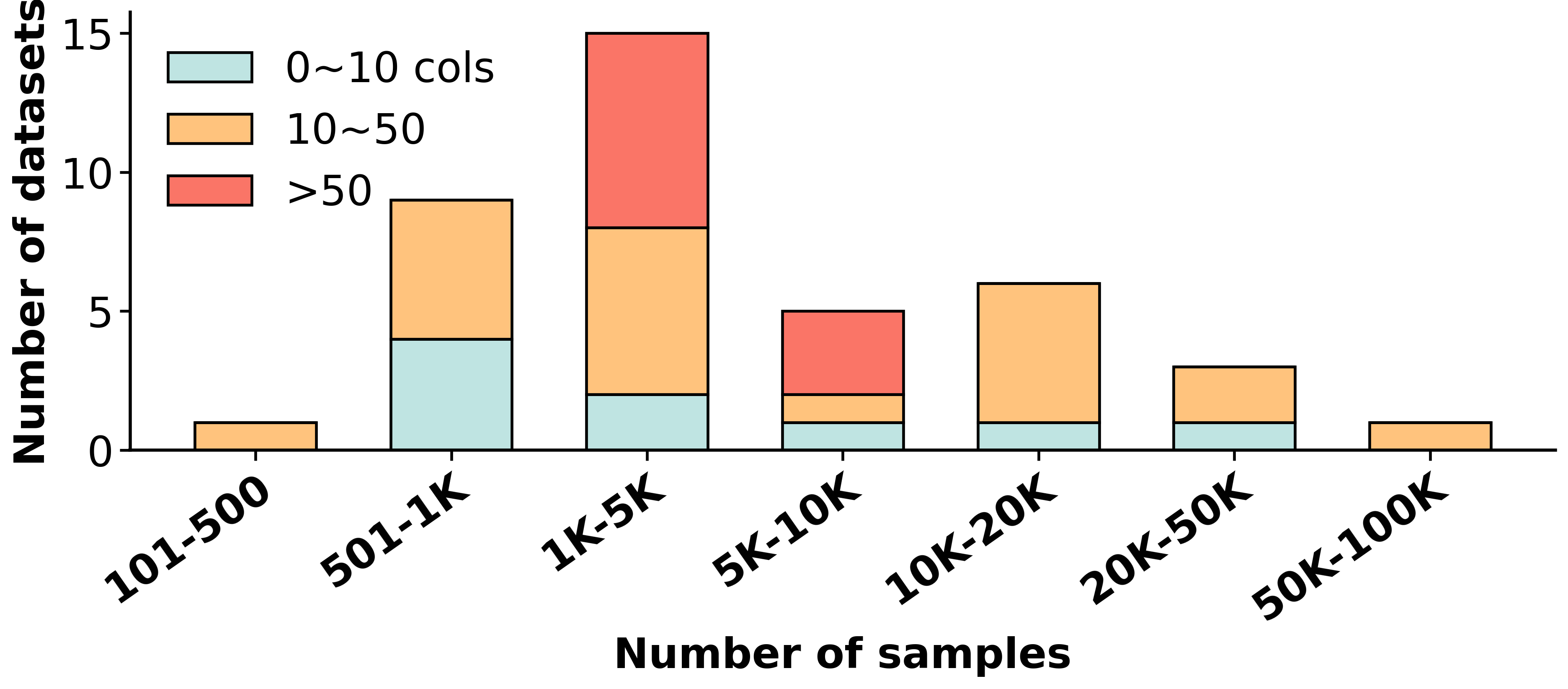}
\label{fig:four_b}}
\caption{Summary of the 40 tabular datasets used in this study. (a) Ratio of numerical and categorical features (b) Number of features and sample distributions.}
\label{fig:dataset}
\end{figure}

\begin{table}[t]
    \centering
    \caption{Number of tabular datasets in which an experimental case yields the best performance at varying numbers of head drops.}
    \begin{tabular}{ccccccccc}
    \toprule 
         Case & Soft  & Head &  \multicolumn{4}{c}{Num. of heads dropped} \\
         & tuning  & dropping strategy & 10 & 20 & 30 & 40 \\
        \midrule
          3 & No  & Random          &  7 & 10 &  6 &  6 \\
          4 & Yes & Random          &  2 &  1 &  0 &  1 \\
        \midrule
          5 & No  & $I_{min}$ $\rightarrow$ $I_{max}$ & 23 & 17 & 22 & 19 \\
          6 & Yes & $I_{min}$ $\rightarrow$ $I_{max}$ &  5 & 10 &  9 & 11 \\
          7 & No  & $I_{max}$ $\rightarrow$ $I_{min}$ &  1 &  1 &  2 &  2 \\
          8 & Yes & $I_{max}$ $\rightarrow$ $I_{min}$ &  2 &  1 &  1 &  1 \\
          \midrule
          &&Total datasets&40&40&40&40\\
        \bottomrule
    \end{tabular}
    \label{tab:result}
\end{table}

\begin{figure} [h]
    \centering
    \includegraphics[width=\linewidth]{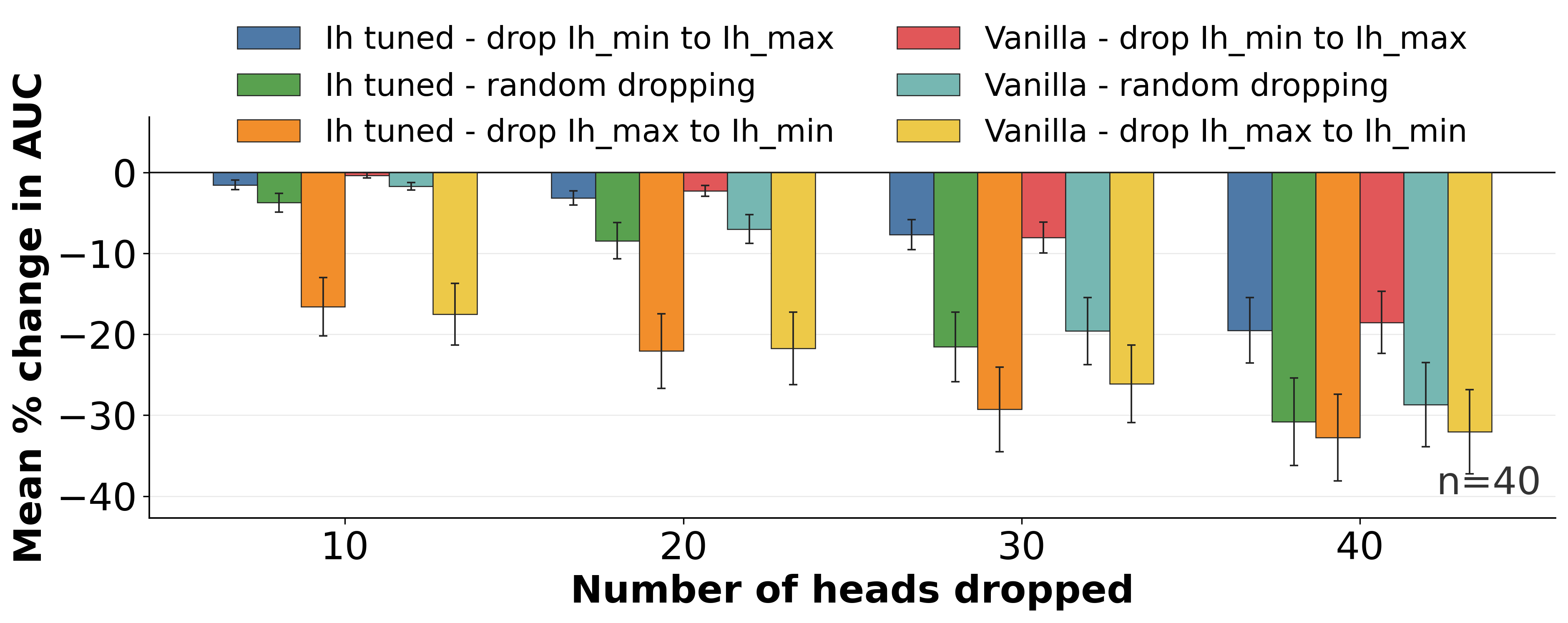}
    \caption{ Mean percentage change in AUC relative to the fixed reference (Case 1: no tuning, no dropping). The changes are averaged across 40 tabular datasets, at 10, 20, 30, and 40 cumulative heads dropped. Error bars denote one standard deviation.}
    \label{fig:barplot}
\end{figure}

\section{Results} \label{results}


\subsection {Tabular datasets}We investigate the impact of data heterogeneity on outcomes by conducting experiments on 40 tabular datasets obtained from the OpenML~\citep{OpenML2013} repository. Due to space limitations, we include the OpenML IDs for individual datasets in the source code repository.  Figure~\ref{fig:dataset} summarizes these datasets with varying size and dimensionality, with sample sizes ranging from 500 to 96,320 and feature counts ranging from 4 to 256 columns. Across all datasets, there are 1,644 features in total, consisting of 1,459 numerical features (88.75\%) and 185 categorical or text-based features (11.25\%), as illustrated in Figure~\ref{fig:four_a}. Figure~\ref{fig:four_b} shows that most datasets contain between 1K and 5K samples, and fewer datasets include larger sample sizes (5K to 100K). 

\begin{figure}[t]
\centering
\subfigure[When $I_h$ tuning of activations during training is the most robust to head dropping. Using the \texttt{pendigits} data set.]{
\includegraphics[width=0.48\textwidth]{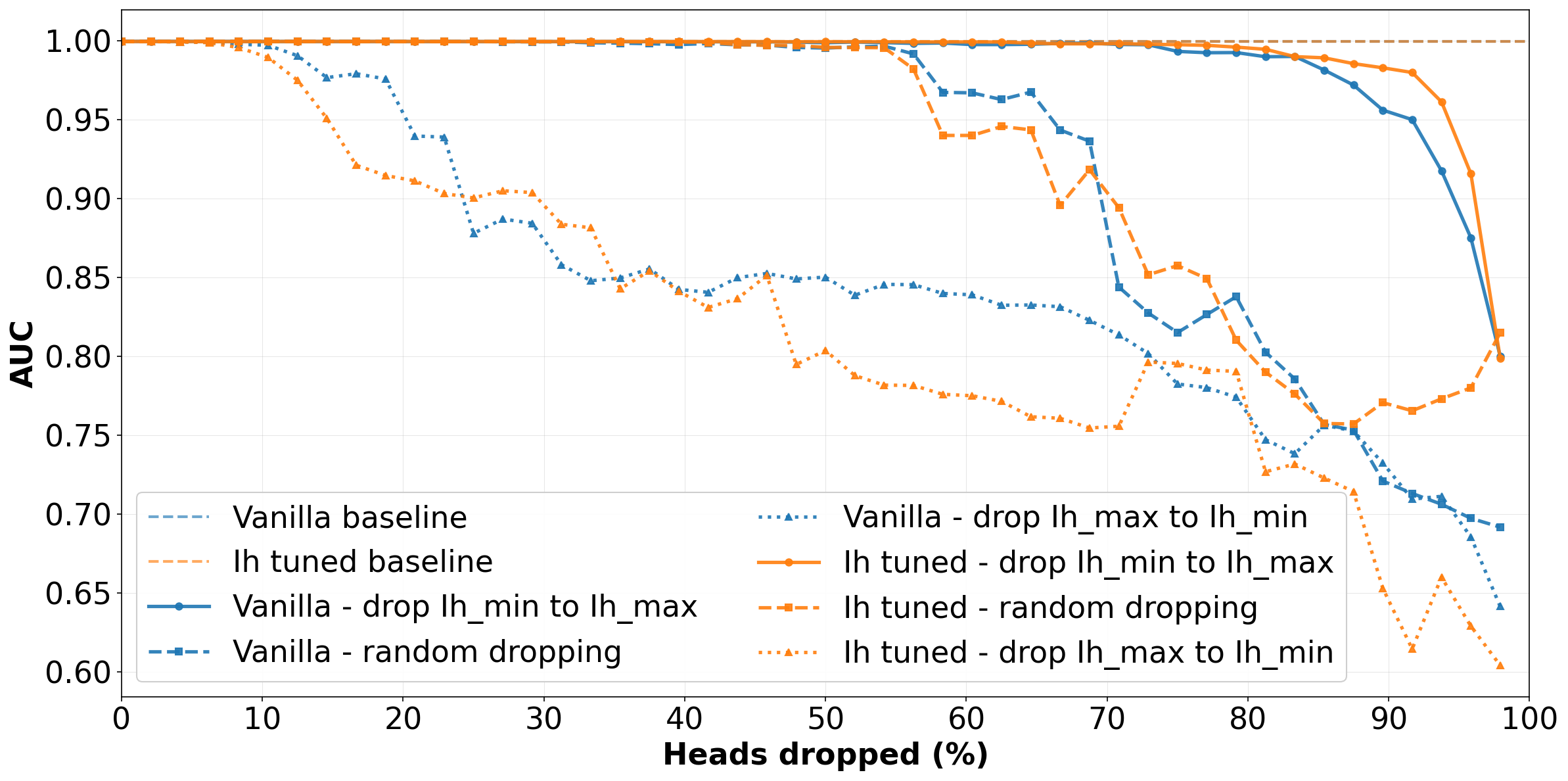}
\label{fig:auc_pendigits}
}
\subfigure[When the vanilla transformer model without activation tuning is the most robust to head dropping. Using the \texttt{eye\_movements} data set.]{
\includegraphics[width=0.48\textwidth]{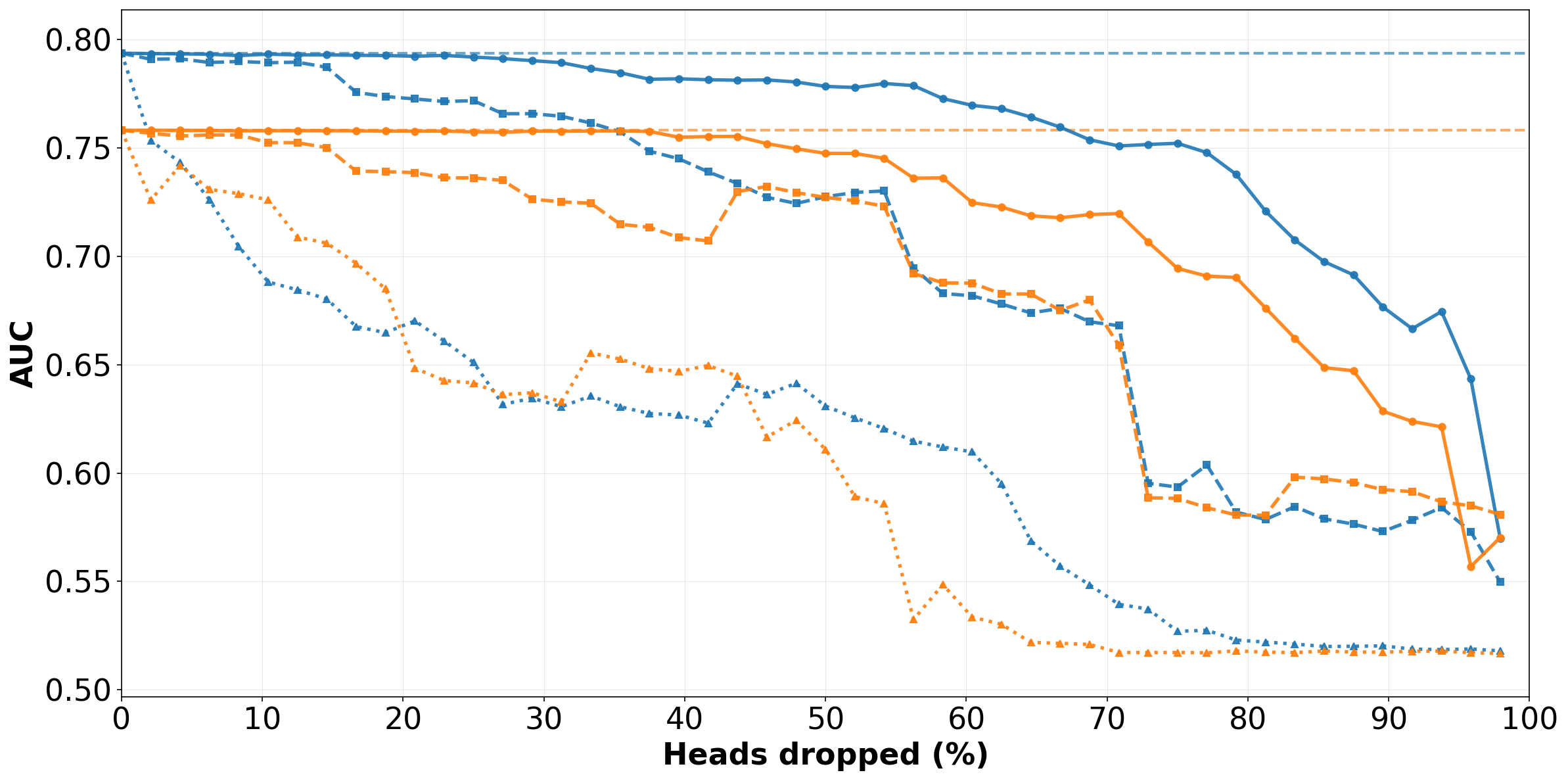}
\label{fig:auc_eye_movements}
}
\vspace{0.25cm}
\subfigure[$I_h$ tuning of activation during training underperforms up to dropping 50\% of heads but stays robust afterward compared to no tuning. Using the \texttt{eucalyptus} data set.]{
\includegraphics[width=0.48\textwidth]{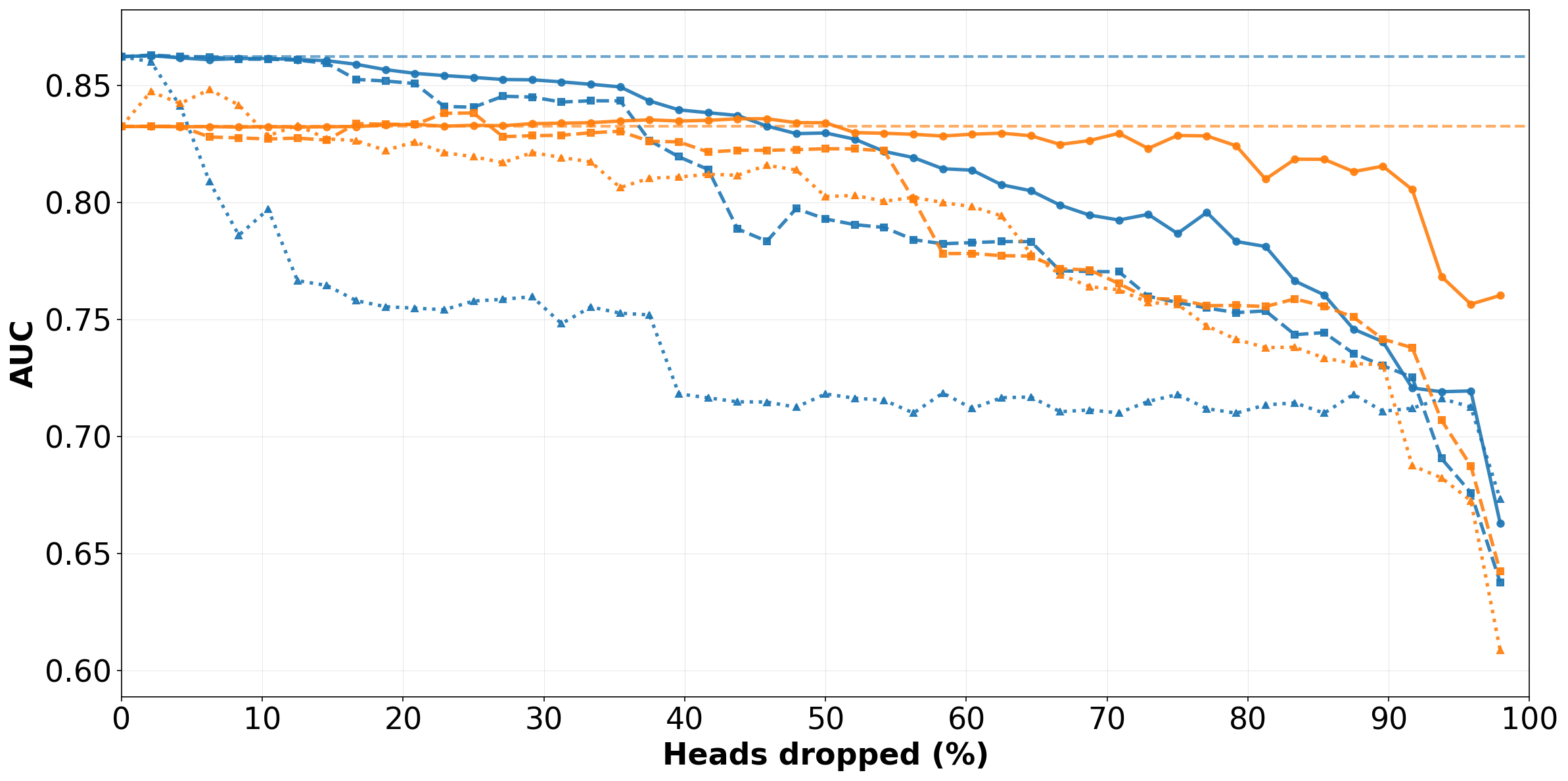}
\label{fig:auc_eucalyptus}
}
\caption{Effects of dropping heads on AUC scores under different experimental cases. 
}
\label{fig:auc_head_dropping}
\vspace{-20pt}
\end{figure}

\subsection{Overall effects of head drop}
Table~\ref{tab:result} highlights how various tabular datasets respond differently in different experimental scenarios. Forty datasets are evaluated at four cumulative head-drop levels (10, 20, 30, and 40), resulting in a total of 160 examples. Among the 160 examples, the strategy of gradually removing heads starting from the lowest $I_h$ values and progressing to higher $I_h$ values yields the most robust performance, as demonstrated by 116 examples (72.5\% of all examples) combined from cases 5 and 6 in Table~\ref{tab:result}. This outcome indicates that the $I_h$ metric effectively identifies the heads that can be dropped with the least effect on the final classification task. The paired Wilcoxon signed-rank test confirms that dropping the least important heads first ($I_{min}$ $\rightarrow$ $I_{max}$) significantly (p$<$0.05) outperforms both random dropping and dropping the most important heads first ($I_{max}$ $\rightarrow$ $I_{min}$)  across all head-drop levels (Cohen's $d$ = 0.40 to 0.80). The effectiveness of head importance score is substantiated when dropping the most important head yields significantly worse (p$<$0.05) performance than random dropping. Among the 116 examples, 35 (30\%) show improved performance when head activations are subject to the proposed soft tuning during model training. The remaining 70\% of the 116 examples do not benefit from soft tuning of head activations. In general, the heterogeneity of the tabular data schema, domain, and data types may play a major role in reaping the benefits of different learning, tuning, and dropping approaches. For example, 29 of the 160 examples (18\%) show the best resilience when heads are dropped randomly, without using $I_h$ scoring for head selection.  

Figure~\ref{fig:barplot} shows the mean percentage change in the classification AUC scores for six experimental cases in forty tabular datasets. When the 10 least important heads are dropped based on $I_h$ scores, the vanilla transformer model yields strong resilience without requiring soft tuning of head activations. The severe negative effect of dropping the most important heads (dropping $I_h$ (max) to $I_h$ (min)) is evident in a significantly large drop in AUC scores. Random removal of heads produces a level of resilience that falls between the best case (removing heads in increasing order of their $I_h$ scores) and the worst case (removing them in decreasing order of their $I_h$ scores).

\begin{figure}[t]
\subfigure[When $I_h$ tuning of activations during training and then head dropping is the best approach. Using the \texttt{pendigits} data set. ] { \includegraphics[trim=1 0 0 0cm, width=0.47\textwidth] {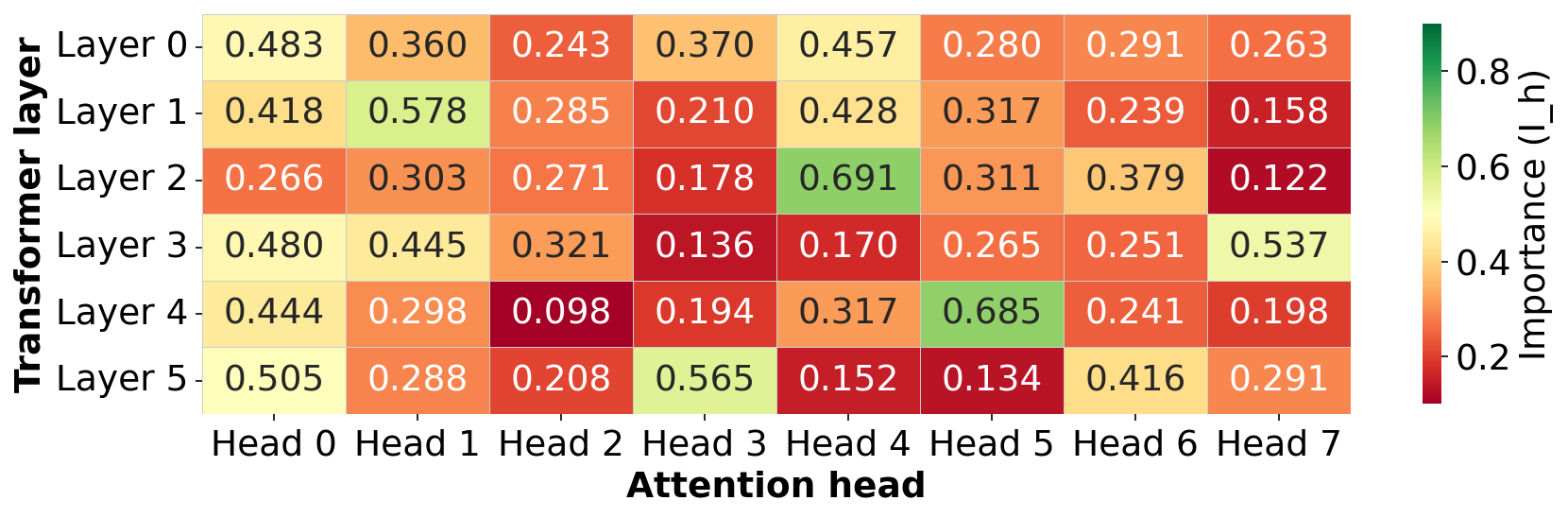}
\label{fig:heat_a}}
\subfigure[When head drop without $I_h$ tuning of activations is the best approach. Using the \texttt{eye\_movements} data set.] { \includegraphics[trim=1 0 0 0cm, clip, width=0.47\textwidth] {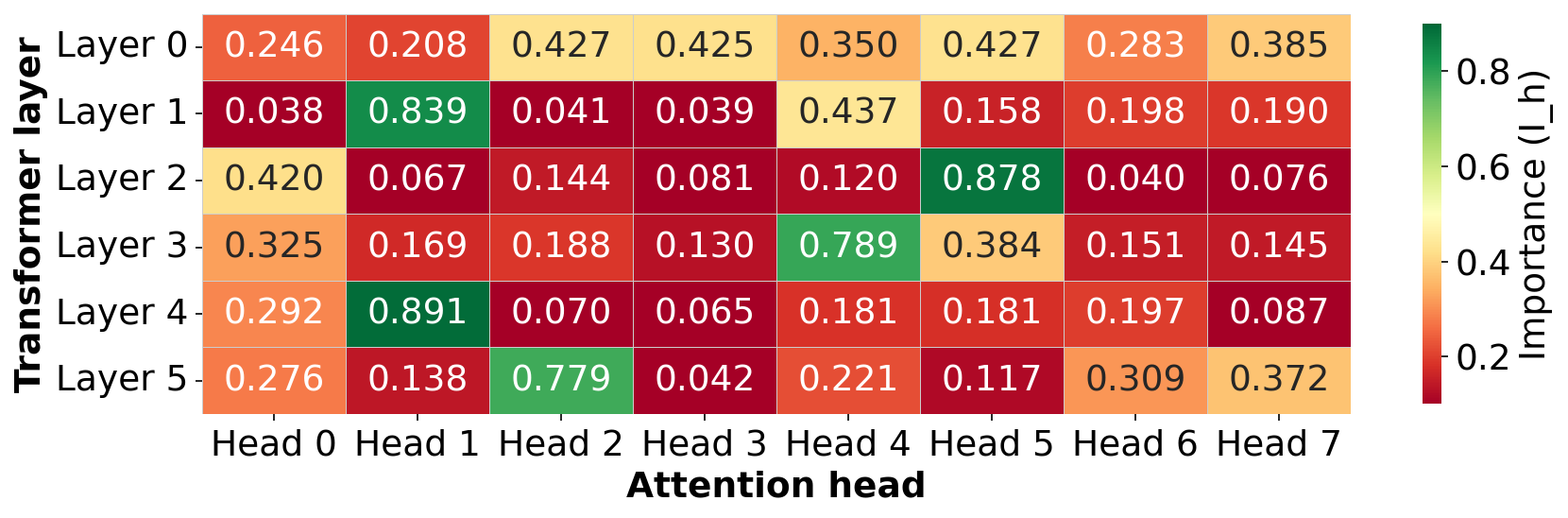}
\label{fig:heat_b}}
\hspace{5pt}
\subfigure[When head drop after $I_h$ tuning outperforms the case without dropping. Using the \texttt{pc3} data set.] { \includegraphics[trim=1 0 0 0cm, clip, width=0.47\textwidth] {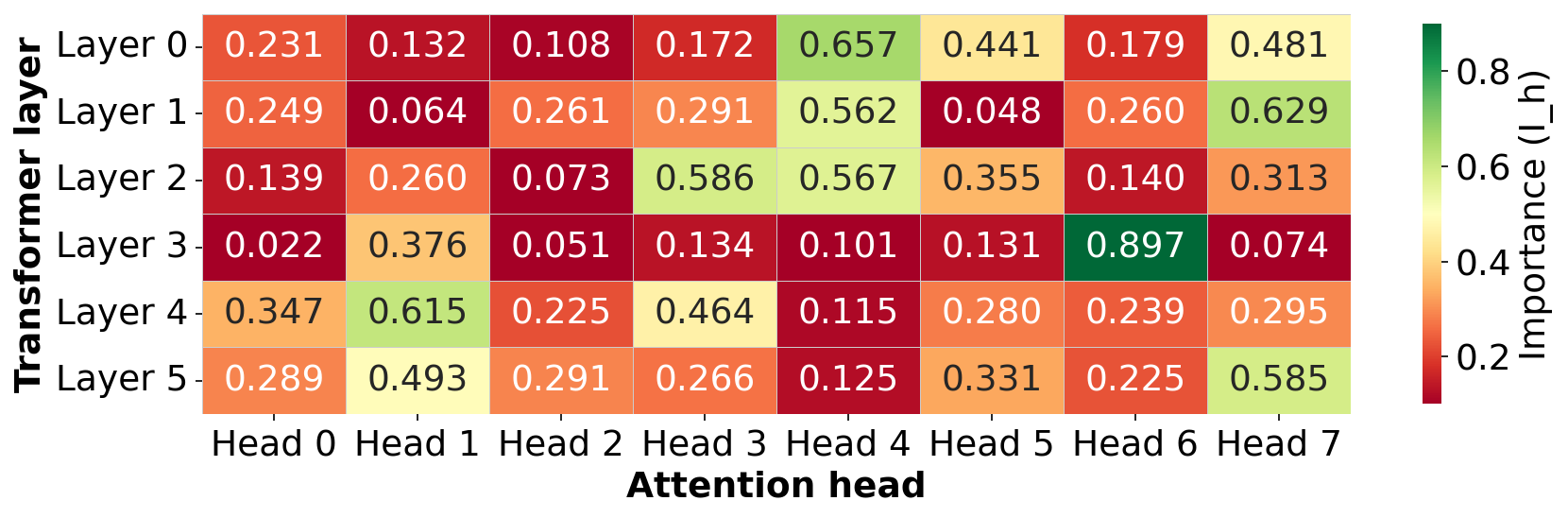}
\label{fig:heat_c}}
\caption{Visualization of attention head importance scores ($I_h$) for three representative learning scenarios.  
}
\label{fig:heat_maps}
\end{figure}

\subsection{Dataset-specific trends}

Figure~\ref{fig:auc_head_dropping} shows the AUC trends with varying percentages of cumulative head drops. The figure illustrates how various tabular datasets exhibit differing degrees of robustness to the head-dropping strategy. The dashed blue and orange lines represent fixed references from case 1 and case 2, respectively. The gap between the two dashed lines indicates that soft tuning of head activations during model training can degrade final classification performance on certain tabular datasets.

Figure~\ref{fig:auc_pendigits} presents an example using the Pendigits data set in which the soft tuning of head activations during training improves the robustness of the model to subsequent head drops. The Pendigits data set encodes handwritten digit strokes as spatially ordered coordinate sequences, which we speculate may benefit from soft tuning of the attention head activations. However, Figure~\ref{fig:auc_eye_movements} shows that soft tuning of head activations has a negative effect while training on the eye-movements dataset. In this example, soft tuning degrades the baseline model performance (case 2) compared to the case without tuning (case 1). The eye-movements dataset consists of sequential, non-image-based reading measurements, which may not be well suited for soft tuning of the attention heads. Figure~\ref{fig:auc_eucalyptus} using the eucalyptus dataset shows similar baseline drop in performance. However, the soft-tuned model maintains its reference AUC scores even after dropping 60\% of the total attention heads, whereas the vanilla model without soft tuning starts to lose performance after dropping 15 heads. The eucalyptus data set is small (736 samples) and has mixed feature types, so soft tuning of head activations may compensate for overfitting due to limited sample size.

\subsection{Visualization of head importance distribution}
Vision models show that knowledge or data representations are hierarchically distributed across deep layers. Figure~\ref{fig:heat_maps} illustrates the attention head importance score distributions for different datasets in different experimental cases. Attention heads with high importance scores are generally scattered throughout the six layers, without showing distinct layer-specific trends. Figure~\ref{fig:heat_a} illustrates that when the attention heads are soft-tuned during training on the pendigit datasets, the importance scores concentrate in the first heads of each attention layer. Several heads with top importance scores appear in the middle heads and attention layers. Compared to Figure~\ref{fig:heat_a}, importance scores for the eye movement data in Figure~\ref{fig:heat_b} show most heads with low importance scores. Several heads with very high importance scores scatter around the mid region of attention layers and head distributions. Variation in head importance patterns may stem from soft tuning in head activations, which could promote a more uniform distribution of head importance across attention layers. In some datasets, the first attention layer contains more important heads than the top layer, though this ordering does not hold consistently across datasets. Figure~\ref{fig:heat_c} illustrates the importance score distribution when dropping heads gives better performance than the vanilla model using the PC3 dataset. In this figure, the lower portion of head 4 (attention layers 0–2) shows high importance scores, in sharp contrast to the upper portion of the attention layers. The opposite is observed for Head 1, where the top half of the attention layers contains the most important heads. Layer-wise, layer 3 has the concentration of the least important heads, except one head that is the most important (score: 0.897) of all heads combined. The PC3 data set consists of software complexity metrics that are mathematically derived from source code. Many of these metrics are inherently correlated, which we speculate may contribute to the concentration of importance in isolated heads.


\subsection{Summary of results}

This paper presents one of the first efforts to study the importance of attention heads in the multi-head transformer architecture while learning tabular datasets. The key findings of this paper can be summarized as follows. First, tabular datasets vary greatly across domains and schemas, and their behavior when processed by multi-head transformers is likewise highly diverse. Therefore, a general pattern or hierarchy observed in vision and language models may not apply to tabular data. Second, the $I_h$ metric is effective in identifying the attention heads that are important for classification tasks that involve tabular datasets. Therefore, head importance scores can be used to measure redundancy in the model, thus helping to interpretability and model compression. Third, the locations of important attention heads vary across datasets, with no consistent concentration in any particular attention layer.

\section{Conclusions} \label{conclu}

This paper develops and evaluates an importance score to interpret how individual attention heads contribute to learning tabular data in multi-head transformer models. The impact of individual heads differs substantially across datasets, indicating that attention heads do not all contribute equally and that attention layers lack any consistent ordering or hierarchy in terms of head-level contributions. One limitation of our work is that all experiments use a single tabular transformer architecture, TransTab. Therefore, gaining insight into how attention heads operate on a particular dataset across different transformer architectures is essential for making effective use of transformer models.

\bibliographystyle{elsarticle-num}
\bibliography{NewPruning}

\end{document}